\documentclass[numbers,final]{diffuspace}

\def\1{\bm{1}}

\DeclareMathAlphabet{\mathsfit}{\encodingdefault}{\sfdefault}{m}{sl}
\SetMathAlphabet{\mathsfit}{bold}{\encodingdefault}{\sfdefault}{bx}{n}

\newcommand{\clover}{Clover-35B-A3B-Preview\xspace}

\RenewDocumentEnvironment{takeaway}{O{Takeaway} +b}
{%
  \refstepcounter{takeaway}%
  \begin{dscallout}[unbreakable]
    \textbf{#1~\thetakeaway.}\ \ignorespaces#2
  \end{dscallout}%
}
{}

\newlength{\benchmarkcolunit}
\title{Clover-35B-A3B-Preview Technical Report}
\shorttitle{Clover-35B-A3B-Preview}
\reporttype{Technical Report}
\date{November 2026}
\author{DiffuSpace Technical Team}
\affiliation{\href{https://diffuspace.ai}{diffuspace.ai}}
\AtBeginEnvironment{thebibliography}{\interlinepenalty=10000}
\begin{document}
\maketitle
\begin{abstract}
We introduce \clover (Clover), a mixture-of-experts diffusion language model with
35 billion total parameters and 3 billion activated parameters, supporting
thinking-mode generation. Clover adapts Gated Delta Networks to block diffusion
by combining recurrent memory of previously generated context with
bidirectional softmax attention inside the current block. Historical memory
readout competes with current-block tokens in a shared softmax, with learnable
block-size calibration and a separate, zero-initialized right-context
convolution branch. Clover leads the seven evaluated agent metrics, including
73.6 on SWE-bench Verified and 51.7 on Terminal-Bench 2.1. On internal agent
workloads, it reaches a reported generation speed of 680 tokens/s using one NVIDIA GB300 GPU at BF16
precision and batch size one. We organize the evaluation around model quality,
drafting and verification costs, and deployment under constrained memory.
We also describe a decision-model variant that generates answers to related
questions in one denoising pass. On an internal set of 500 groups, joint
prediction raises question accuracy from 90.37\% to 93.88\% and all-correct
group accuracy from 65.4\% to 77.0\%.
\begin{resources}
  \resource{\githubicon}{Code}{https://github.com/DiffuSpace/Clover-35B-A3B-Preview}
  \resource{\hficon}{Weights}{https://huggingface.co/Diffuspace/Clover-35B-A3B-Preview}
\end{resources}
\end{abstract}

\section{Introduction}
\label{sec:introduction}
Diffusion language models (DLMs) offer a different route to language generation:
reconstructing sequences through iterative denoising, with multiple tokens
predicted in parallel. This formulation connects flexible generation order
with the prospect of efficient inference. Progress in diffusion formulations,
model scaling, and decoding has brought DLMs from specialized text-generation
settings to increasingly capable general-purpose models. The next step is to
translate these advances into strong performance and practical efficiency
across demanding applications.

Continuous text diffusion models established generation through denoising
learned token representations. Diffusion-LM explored controllable generation,
while DiffuSeq and DiffuSeq-v2 developed sequence-to-sequence diffusion
and a bridge between discrete corruption and continuous
representations~\citep{li2022diffusionlmimproves,gong2023diffuseq,gong2023diffuseqv2}.
Alongside these models, D3PM formalized structured categorical
corruption~\citep{austin2021structureddenoisingdiffusion}. Reparameterized
Discrete Diffusion Models (RDM) exposed a routing interpretation of backward
sampling~\citep{zheng2023reparameterizeddiscretediffusion}; SEDD and MDLM further
developed discrete score modeling and masked-token
reconstruction~\citep{lou2023discretediffusionmodeling,sahoo2024simpleeffectivemasked}.

Beyond text reconstruction, Diffusion of Thought explored denoising explicit
reasoning trajectories~\citep{ye2024diffusionofthought}. \citet{ye2025beyondautoregression} investigated discrete diffusion for planning and constraint-solving tasks,
using multi-granularity learning and adaptive generation
order. These studies established
capability directions that complement the development of general-purpose DLMs.

Scaling made these ideas accessible through larger models. Adapting
pretrained autoregressive checkpoints provided one route to diffusion
language modeling~\citep{gong2025scalingdlm}; LLaDA and Dream 7B broadened the
publicly available model ecosystem~\citep{nie2025llmdiff,ye2025dream}, and
code-focused models such as Dream-Coder 7B extended masked diffusion to code
generation and reinforcement learning with verifiable
rewards~\citep{xie2025dreamcoder,gong2026diffucoder}. Block diffusion combines sequential
progress across blocks with parallel denoising inside each
block~\citep{arriola2025blockdiffusion}. Fast-dLLM and Fast-dLLM v2 developed
complementary inference and adaptation techniques based on caching and
confidence-aware parallel decoding~\citep{wu2025fastdllmtrainingfreeaccelerationdiffusion,wu2025fastdllmv2}. DreamReasoner further extends block diffusion to long chain-of-thought reasoning through block-size curriculum learning~\citep{wu2026dreamreasoner}.

Industrial development now spans commercial systems such as Mercury and
publicly released models such as DiffusionGemma and
LLaDA2.2~\citep{inceptionlabs2025mercury,team2026diffusiongemmatechnicalreport,inclusionai2026llada22}.
These systems have expanded the scope and deployment options of diffusion
language modeling.

As the model ecosystem matures, agentic applications provide a demanding test
of this progress: they require both capable tool use and efficient generation
through repeated interactions with an environment. Strong results on isolated
reasoning or code-generation tasks do not by themselves establish reliable
repository-level problem solving. Competitive agent performance remains a challenge for open DLMs. Recent models
already include agent evaluations, but the reproduced
comparison shows that LLaDA2.2 flash and DiffusionGemma still trail the evaluated
Qwen baselines on both Terminal-Bench 2.1~\citep{merrill2026terminalbench} and SWE-bench Verified~\citep{chowdhury2024swebenchverified}
(Table~\ref{tab:main-results}). Advancing agent capability while retaining efficient
parallel generation is a central motivation for Clover.

To this end, we present \clover, an efficient and capable diffusion language model designed for agentic applications. It combines a 35B-total, 3B-active mixture-of-experts
architecture with a bidirectional adaptation of Gated Delta Networks.
Completed
context is represented by recurrent memory, while tokens in the active block
interact through softmax attention. A shared softmax combines historical and
block-local information, with learnable block-size calibration and a separate
right-context convolution branch. Our evaluation emphasizes agentic coding
and tool use alongside general capabilities, and examines the cost of turning
parallel predictions into accepted output tokens.

The report first introduces the modeling and decoding background, then
presents Clover's architecture, benchmark results, and runtime analysis.
Separate sections examine on-device deployment and a decision-model variant
that generates related answers jointly.

We release model weights, supervised fine-tuning code, and the inference
engine, with particular emphasis on dedicated infrastructure for running
Clover locally on consumer devices. We will continue to improve inference speed
and model performance, expand multimodal capabilities, and contribute to
the continued development of diffusion language models.

\section{Preliminaries}
\label{sec:preliminaries}
\subsection{Masked diffusion and block generation}
Let $x_0=(x_0^1,\ldots,x_0^L)$ be a clean token sequence and $m$ be the mask token.
In an absorbing-mask forward process~\citep{austin2021structureddenoisingdiffusion}, each position is retained with probability
$\alpha_t$ and replaced by $m$ otherwise:
\begin{equation}
 q(x_t^i\mid x_0^i)=\alpha_t\,\delta_{x_0^i}(x_t^i)
 +(1-\alpha_t)\,\delta_m(x_t^i),
 \qquad \alpha_0=1,\quad \alpha_1=0.
 \label{eq:masking}
\end{equation}
Here $\delta$ denotes a point mass. A masked diffusion model (MDM) predicts clean
tokens from the corrupted sequence. For a differentiable decreasing schedule,
the standard continuous-time masked reconstruction objective~\citep{sahoo2024simpleeffectivemasked,shi2024simplified} can be written as
\begin{equation}
 \mathcal{L}_{\mathrm{MDM}}=
 \mathbb{E}_{t\sim\mathcal{U}(0,1),\,x_0,\,x_t}
 \left[w(t)\sum_{i:x_t^i=m}\bigl(-\log p_\theta(x_0^i\mid x_t,t)\bigr)\right],
 \qquad w(t)=-\frac{\alpha_t'}{1-\alpha_t}.
 \label{eq:mdm}
\end{equation}

Fast-dLLM accelerates
masked diffusion through approximate KV reuse and confidence-aware parallel
decoding, while Fast-dLLM v2 develops block-level caching~\citep{wu2025fastdllmtrainingfreeaccelerationdiffusion,wu2025fastdllmv2}. In block diffusion~\citep{arriola2025blockdiffusion}, completed blocks supply context for
subsequent blocks. We divide the output into blocks $B_1,\ldots,B_K$ and let
$c_k$ contain the prompt and completed blocks preceding $B_k$. The model
denoises the current block conditional on $c_k$, with bidirectional access to
its partially reconstructed contents. 
Block size, the number of
denoising passes, and the token selection rule together determine how much
parallelism is exposed. See Section~\ref{sec:training} for Clover's training setup.

\subsection{Self-speculative decoding}
\label{sec:selfspec-background}
Speculative decoding accelerates autoregressive generation by verifying
multiple draft tokens in a single target-model pass~\citep{cai2024medusa,li2025eagle3}.
Recently, diffusion language models offer a natural drafting mechanism through
parallel block prediction, as demonstrated by
DFlash~\citep{chen2026dflash} and DFlash 2~\citep{incoai2026dflash2}.
Self-speculative decoding integrates
these roles within the same model: diffusion-style block prediction supplies
candidate tokens, and autoregressive verification determines which prefix can
be committed. NVIDIA's Nemotron-Labs-Diffusion supports autoregressive,
diffusion, and self-speculation decoding within one
model~\citep{fu2026nemotronlabsdiffusion}. 


\subsection{Gated Delta Networks}
\label{sec:gdn-background}
Gated Delta Networks combine recurrent state decay with a delta-rule memory
update~\citep{yang2025gateddelta}. With column-vector keys $k_i$, values $v_i$,
and queries $q_i$, one conventional orientation of the recurrence is
\begin{align}
 \widetilde S_i &= \gamma_i S_{i-1},\\
 S_i &= \widetilde S_i + \beta_i
        (v_i-\widetilde S_i k_i)k_i^\top,\qquad
 o_i=S_iq_i.
 \label{eq:gdn}
\end{align}
Here $S_i\in\mathbb{R}^{d_v\times d_k}$ stores the recurrent memory, while
$\gamma_i$ and $\beta_i$ control retention and the update. The delta rule writes
the residual between a new value and the value retrieved with its key.
Unlike a token-wise attention cache, this state has a fixed shape for a fixed
head configuration. \eqref{eq:gdn} suppresses head indices, feature
normalization, and surrounding projections.

A causal recurrence summarizes information to the left of a token. Clover's
adaptation retains this role for completed context and introduces bidirectional
interactions for tokens being denoised together. The distinction is between
\emph{completed history} and the \emph{active block}; bidirectionality does not
expose unseen future blocks.

\section{Clover-35B-A3B-Preview: Architecture and Evaluation}
\label{sec:clover}

\subsection{Model overview}
\label{sec:model-overview}
\begin{figure}[htbp]
 \centering
 \includegraphics[width=\linewidth]{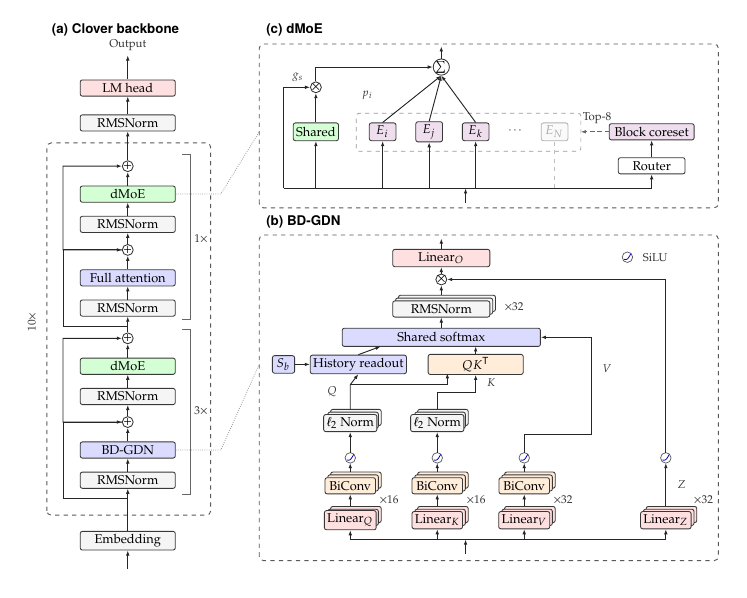}
 \caption{\textbf{Clover architecture.}
 (a) Three BD-GDN--dMoE layers and one full-attention--dMoE layer form a group
 repeated ten times. Backbone normalizations are RMSNorm.
 (b) Pink boxes are linear projections; the $Q$, $K$, $V$, and $Z$ paths carry
 query, key, value, and output-gate features. Stacked boxes indicate multiple
 heads: 16 query heads, 16 key heads, and 32 value heads; each query/key head
 serves two value heads. The historical readout from $S_b$ and block-local
 information are combined by the shared softmax. Output RMSNorm operates per
 value head, and gated head outputs are concatenated before the output projection.
 (c) dMoE combines token-level top-eight routing within a block coreset with
 a shared expert. Solid lines carry features, dashed arrows show routing,
 and gray experts are inactive. For one token, $p_i$ is the normalized weight
 of routed expert $i$, and $g_s$ is the shared expert's sigmoid gate.}
 \label{fig:architecture}
\end{figure}

\clover adapts the text backbone of Qwen3.6-35B-A3B~\citep{qwen36_35b_a3b} into a block diffusion
language model with approximately 35B total parameters and 3B activated
parameters per token. Its hybrid backbone combines recurrent sequence
mixing, full attention, and sparse expert computation. We adapt two
components to block denoising: \emph{Block-diffusion Gated DeltaNet}
(BD-GDN) combines compressed clean history with bidirectional interactions
inside the current block, while \emph{dMoE} coordinates expert selection
across that block. Figure~\ref{fig:architecture} summarizes the architecture.

\paragraph{Hybrid backbone.}
The model has 40 decoder layers with hidden width $D=2{,}048$. A group of
three BD-GDN layers followed by one full-attention layer is repeated ten
times, giving 30 recurrent mixers and 10 full-attention mixers. Each mixer
is followed by a dMoE~\citep{feng2026dmoe} module, with pre-normalization and residual connections:
\begin{equation}
 \begin{aligned}
 U^\ell &= H^\ell+\operatorname{Mixer}_\ell
                  (\operatorname{RMSNorm}(H^\ell)),\\
 H^{\ell+1} &= U^\ell+\operatorname{dMoE}_\ell
                  (\operatorname{RMSNorm}(U^\ell)).
 \end{aligned}
 \label{eq:clover-layer}
\end{equation}
Here $\ell$ indexes decoder layers, $H^\ell\in\mathbb R^{L\times D}$ is
the input hidden-state sequence of length $L$, and $U^\ell$ is the
intermediate state after sequence mixing. $\operatorname{Mixer}_\ell$
is either BD-GDN or full attention; RMSNorm~\citep{zhang2019rmsnorm} normalizes each token over
its feature dimension with a learned feature-wise scale.

BD-GDN uses $H_k=16$ query/key heads and $H_v=32$ value heads, with
key and value dimensions $d_k=d_v=128$. Each query/key head serves two
value heads. Full-attention layers use 16 query heads, two K/V heads,
and head dimension 256, with query/key normalization, partial rotary
position embeddings~\citep{su2021roformer}, and a sigmoid output gate. The model supports
131,072 tokens across the prompt and generated response, with untied
input/output embeddings and a vocabulary of 248,320 entries.

\begin{takeaway}
Our architectural exploration led us to prioritize bidirectional interaction
within Clover's current active block. By enabling tokens to use context from both
directions as fully as possible, we aim to preserve and fully exploit the
bidirectional modeling capability of diffusion, while retaining efficient
access to the completed prefix.
\end{takeaway}

\subsection{Block-diffusion Gated DeltaNet}
\label{sec:architecture}
BD-GDN separates \emph{writing completed context} from \emph{reading it
during denoising}. The clean path retains the gated delta-rule recurrence
of Section~\ref{sec:gdn-background}~\citep{yang2025gateddelta}. The noisy path reads the resulting
prefix state and combines it with bidirectional attention over the current
block. This prefix state stays fixed throughout the block's denoising passes.

\paragraph{Projected features and notation.}
Let $h_i\in\mathbb R^D$ be the RMS-normalized input to the BD-GDN at token
position $i$. Linear projections produce query, key, and value features;
these pass through the short convolution and SiLU activation described
below. The resulting per-head vectors are $q_i,k_i\in\mathbb R^{d_k}$
and $v_i\in\mathbb R^{d_v}$. Queries and keys then undergo
\begin{equation}
 \widehat q_i=\frac{q_i}{\sqrt{\lVert q_i\rVert_2^2+\epsilon}},\qquad
 \widehat k_i=\frac{k_i}{\sqrt{\lVert k_i\rVert_2^2+\epsilon}},
 \qquad \epsilon=10^{-6}.
 \label{eq:clover-qk-norm}
\end{equation}
Here $\lVert\cdot\rVert_2$ is the Euclidean norm over one head's features;
the hats denote the normalized vectors. We describe one value head below,
omitting layer and head indices on its vectors and recurrent state.
Query/key heads are repeated to match the 32 value heads.

\paragraph{Clean memory update.}
For a clean token, the learned retention and write gates are
\begin{equation}
 \alpha_i=\exp\!\left[-\exp(a_{\log})
       \operatorname{softplus}(w_a^\top h_i+b_\Delta)\right],\qquad
 \beta_i=\sigma(w_\beta^\top h_i).
 \label{eq:clover-gates}
\end{equation}
The vectors $w_a,w_\beta\in\mathbb R^D$ project the token to scalar gates;
$a_{\log}$ and $b_\Delta$ are learned scalars for this head. We use
$\sigma(u)=1/(1+e^{-u})$ and $\operatorname{softplus}(u)=\log(1+e^u)$.
Thus $\alpha_i$ controls retention of the previous memory and $\beta_i$
controls the strength of the new write. The update is
\begin{equation}
 \begin{aligned}
 \widetilde S_i &= \alpha_i S_{i-1},\\
 S_i &= \widetilde S_i+
       \beta_i(v_i-\widetilde S_i\widehat k_i)\widehat k_i^\top,\\
 o_i^{\mathrm{clean}}&=S_i\widehat q_i/\sqrt{d_k}.
 \end{aligned}
 \label{eq:clover-clean}
\end{equation}
The memory $S_i\in\mathbb R^{d_v\times d_k}$ stores completed context,
and $\widetilde S_i$ is its decayed value before the write. The residual
$v_i-\widetilde S_i\widehat k_i$ corrects the value retrieved with the
current key; $o_i^{\mathrm{clean}}\in\mathbb R^{d_v}$ is the clean output.
Each packed example starts from a zero state; padding neither writes nor decays the memory.

\paragraph{Combining history and the current block.}
Let $b$ index the block being denoised, $S_b$ denote its clean prefix state,
and $\mathcal A_b$ be the set of valid token positions in that block.
For query position $j\in\mathcal A_b$, the historical readout is
\begin{equation}
 r_j=\frac{S_b\widehat q_j}{\sqrt{d_k}},\qquad
 r_j\in\mathbb R^{d_v}.
 \label{eq:clover-history}
\end{equation}
Here $S_b$ is the state at the \emph{block boundary}, whereas $S_i$ in
\eqref{eq:clover-clean} is indexed by a clean token. The readout retains
the $1/\sqrt{d_k}$ scaling of the clean GDN output. It does not apply the
noisy query's decay gate: $\alpha$ and $\beta$ govern clean memory updates.

BD-GDN treats $r_j$ as one information unit alongside the block's token
values. For each layer and value head $h$, we learn a local logit scale
$s_h$, a history bias $b_h$, and a block-size coefficient $c_h$. With
$N_b=|\mathcal A_b|$ and reference size $N_{\mathrm{ref}}=32$, the logits are
\begin{equation}
 \ell_{ji}=s_h\widehat q_j^\top\widehat k_i\quad(i\in\mathcal A_b),
 \qquad
 \ell_{j,\mathrm{hist}}=b_h+c_h\log(N_b/N_{\mathrm{ref}}).
 \label{eq:clover-logits}
\end{equation}
The index $i$ runs over current-block keys, while $j$ identifies the query.
The subscript $h$ indexes value heads, not token positions. One softmax
normalizes the $N_b$ local logits and the history logit together:
\begin{equation}
 \begin{aligned}
 (\{p_{ji}\}_{i\in\mathcal A_b},p_{j,\mathrm{hist}})
     &=\operatorname{softmax}
        (\{\ell_{ji}\}_{i\in\mathcal A_b},\ell_{j,\mathrm{hist}}),\\
 o_j &=p_{j,\mathrm{hist}}r_j+
                  \sum_{i\in\mathcal A_b}p_{ji}v_i.
 \end{aligned}
 \label{eq:clover-bdgdn}
\end{equation}
The weights $p_{ji}$ and $p_{j,\mathrm{hist}}$ sum to one for each query
and head; $o_j\in\mathbb R^{d_v}$ is their weighted output. A query can
therefore use both the compressed prefix and individual current-block
tokens. Local logits use the learned scale $s_h$ on normalized Q/K;
the $1/\sqrt{d_k}$ factor scales the historical \emph{value}, not these logits.
All valid block positions participate, including the query itself and
unresolved mask tokens. Padding is excluded from $\mathcal A_b$.

\paragraph{Learnable block-size calibration.}
History occupies one softmax entry, while the number of local entries
grows with block size. The term $c_h\log(N_b/N_{\mathrm{ref}})$ compensates
for this change in the number of competitors. To see the effect, suppose
all local logits equal a common value $\mu$ and $c_h=1$. Then
\begin{equation}
 p_{j,\mathrm{hist}}=
 \frac{\exp(b_h)}{\exp(b_h)+N_{\mathrm{ref}}\exp(\mu)},
 \label{eq:clover-calibration}
\end{equation}
which is independent of $N_b$. Learning $c_h$ lets each head adjust this
compensation as its local logit distribution changes. We initialize
$(s_h,b_h,c_h)=(10,5,1)$. The count $N_b$ includes unresolved mask positions
and excludes padding, so partial blocks use their actual valid length.

\begin{figure}[htbp]
 \centering
 \includegraphics[width=\linewidth]{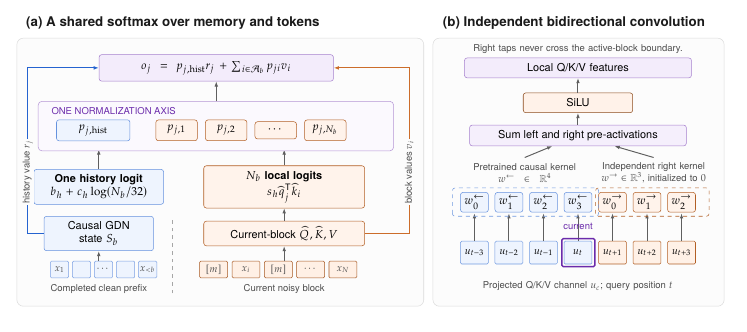}
 \caption{\textbf{The two block-diffusion adaptations in BD-GDN.}
 (a) The query-specific history readout $r_j$ from
 \eqref{eq:clover-history} competes with $N_b$ current-block values in one
 softmax. Its logit includes learnable block-size calibration.
 (b) Independent causal and right-context depthwise kernels are summed
 before SiLU. The right taps start at zero and read only valid successors
 in the current block; clean prefill and commit use the causal kernel.}
 \label{fig:bdgdn-detail}
\end{figure}

\paragraph{Independent bidirectional short convolution.}
The query, key, and value projections form a tensor with
$C=2H_kd_k+H_vd_v=8{,}192$ channels. Let $u_{c,i}$ be channel $c$ of this
tensor at token position $i$, before convolution. We combine a causal
depthwise kernel of width $W=4$ with an independent right-context kernel:
\begin{equation}
 \begin{aligned}
 a_{c,i}&=\sum_{r=0}^{W-1}w^{\leftarrow}_{c,r}
                     u_{c,i-(W-1-r)}
       +\sum_{m=1}^{W-1}w^{\rightarrow}_{c,m}u_{c,i+m},\\
 \bar u_{c,i}&=\operatorname{SiLU}(a_{c,i}).
 \end{aligned}
 \label{eq:clover-conv}
\end{equation}
Here $r$ indexes causal taps, $m$ is the rightward token offset, and
$w^{\leftarrow}$ and $w^{\rightarrow}$ are the two learned kernel weights.
Their outputs are summed before one activation
$\operatorname{SiLU}(u)=u\sigma(u)$; $\bar u$ is then split into Q/K/V.
The convolution is bias-free and operates independently on each channel.
Left inputs come from committed convolution history and earlier positions
of the current block. Right inputs are restricted to valid successors in
that block; missing inputs contribute zero. Thus no right tap crosses a
block or packed-example boundary (Figure~\ref{fig:bdgdn-detail}).

We retain the pretrained causal kernel and initialize all three right taps
to zero. This preserves the causal convolution at initialization while
allowing the added taps to learn local right-context features. Clean prefill
and causal commit use only the causal kernel. Across 30 BD-GDN layers,
the right kernels add 737,280 parameters and the three per-head softmax
scalars add 2,880, for a total of 740,160 new parameters.

\paragraph{Output gating and cache updates.}
Each value head applies RMSNorm over its $d_v$ output features and a SiLU
output gate. Concatenating all heads gives the mixer output
\begin{equation}
 z_j=W_z h_j,\qquad
 y_j=W_o\operatorname{Concat}_h\left[
   \operatorname{RMSNorm}(o_{j,h})\odot\operatorname{SiLU}(z_{j,h})\right].
 \label{eq:clover-output}
\end{equation}
The learned projection $W_z$ produces $H_vd_v$ gate features, split into
per-head vectors $z_{j,h}\in\mathbb R^{d_v}$; $W_o$ maps the concatenated
features back to width $D$. The symbol $\odot$ denotes element-wise
multiplication, and $y_j$ enters the mixer residual in
\eqref{eq:clover-layer}. Every denoising pass reads the same committed
recurrent and convolution caches. Once the block is resolved, causal replay
recomputes its clean activations and advances these caches. Full-attention
preview K/V are likewise temporary and become persistent only through
the clean path.

\begin{samepage}
\subsection{Block-aware expert routing}
\label{sec:dmoe}
Parallel block prediction can spread token-level routing across many
experts. Following dMoE~\citep{feng2026dmoe}, we coordinate these choices by first selecting a shared expert
pool for the block, then routing individual tokens within that pool
(Figure~\ref{fig:architecture}c). Each module contains 256 routed experts
and one shared expert, all using SwiGLU~\citep{shazeer2020glu} with intermediate width 512.
\par\end{samepage}

For the RMS-normalized dMoE input $f_j\in\mathbb R^D$, a learned router
$W_r\in\mathbb R^{256\times D}$ produces probabilities
$a_{je}=\operatorname{softmax}(W_r f_j)_e$ over routed experts
$e\in\{1,\ldots,256\}$. We first identify each valid block token's
top-eight experts and sum their probabilities $a_{je}$ by expert across
the block. Sorting these aggregate scores, we retain the smallest expert
prefix covering a fraction $\tau$ of their total mass. This is the block's
\emph{expert coreset}; the released configuration uses $\tau=0.9$.
Each layer and denoising pass computes its own coreset. Invalid padding
does not contribute to its scores.

We mask router logits outside the coreset and rerun token-level top-eight
selection. Let $\mathcal T_j$ contain token $j$'s selected experts with
nonzero weight, and let
$p_{je}=a_{je}/\sum_{e'\in\mathcal T_j}a_{je'}$ for $e\in\mathcal T_j$.
The module output is
\begin{equation}
 \operatorname{dMoE}(f_j)=g_s(f_j)E_s(f_j)
             +\sum_{e\in\mathcal T_j}p_{je}E_e(f_j),
 \qquad g_s(f_j)=\sigma(w_s^\top f_j).
 \label{eq:clover-dmoe}
\end{equation}
Here $E_e$ is routed expert $e$ and $E_s$ is the shared expert, each mapping
width $D$ back to width $D$; $w_s\in\mathbb R^D$ is the shared gate's learned
projection. The shared expert is always evaluated
and is outside the coreset selection. This block-level restriction
encourages expert reuse across parallel predictions while retaining
token-specific weights. Clean causal processing uses ordinary token-level
routing, so its expert choices do not depend on later tokens.

\subsection{Training recipe}
\label{sec:training}
We extend Megatron-Core~\citep{narayanan2021megatron} and
Megatron Bridge~\citep{nvidia2026megatronbridge} to train the hybrid backbone
with block diffusion. The implementation carries the same clean/noisy
distinction through sequence packing, attention masks, recurrent state,
expert routing, and distributed execution.

\paragraph{Two-stream training and causal isolation.}
For each example $s$, we construct a corrupted copy $x_t^{(s)}$ and a clean
copy $x_0^{(s)}$, where $t$ denotes the corruption level introduced in
Section~\ref{sec:preliminaries}. The copies share semantic token positions
and are stored as
$[x_t^{(1)}\mid x_0^{(1)}\mid x_t^{(2)}\mid x_0^{(2)}\mid\cdots]$.
Clean queries are token-causal. A noisy query can read clean tokens
strictly before its block and noisy tokens within that block, but cannot
read its clean target block or other noisy blocks. Explicit example,
stream, and block identities enforce this visibility after packing.
Padding inserted for alignment neither advances clean memory nor becomes
a supervision target. For each microbatch, we uniformly sample the block
size from $\{4, 8, 16, 32\}$ and use the resulting boundaries consistently
for attention, BD-GDN, right-context convolution, and expert routing.

\paragraph{Denoising and auxiliary causal supervision.}
The denoising objective uses \emph{no-shift} alignment: logits at position
$i$ predict $x_0^i$, with corruption, labels, and loss masks aligned to the
same position. The auxiliary clean-stream objective uses \emph{shifted}
next-token prediction, where position $i$ predicts $x_0^{i+1}$.
For a sampled corruption,
let $\widetilde x_b$ denote noisy block $b$, and $c_b$ its clean prefix,
including the prompt. Let $\mathcal M$ be the corrupted positions eligible for training and
$\mathcal T$ the clean positions with a valid, supervised next-token target
in the same example. The supported joint objective is
\begin{equation}
 \begin{aligned}
 \mathcal L_{\mathrm{BD}}&=
 \frac{1}{|\mathcal M|}\sum_{i\in\mathcal M}w(t_i)
 [-\log p_\theta(x_0^i\mid \widetilde x_{b(i)},c_{b(i)})],\\
 \mathcal L_{\mathrm{AR}}&=
 \frac{1}{|\mathcal T|}\sum_{i\in\mathcal T}
 [-\log p_\theta(x_0^{i+1}\mid x_0^{\leq i})],\\
 \mathcal L&=\lambda_{\mathrm{BD}}\mathcal L_{\mathrm{BD}}
          +\lambda_{\mathrm{AR}}\mathcal L_{\mathrm{AR}}.
 \end{aligned}
 \label{eq:clover-training}
\end{equation}
Here $\theta$ denotes model parameters and $p_\theta$ its predicted token
distribution. The index $i$ runs over semantic token positions, and $b(i)$
identifies the block containing position $i$. The noise level associated
with position $i$ is $t_i$, and $w(t_i)$ is its configured
weight; $w=1$ gives unweighted masked cross-entropy. The coefficients
$\lambda_{\mathrm{BD}}$ and $\lambda_{\mathrm{AR}}$ weight the two losses.
An empty selection contributes zero. In the causal term, $i+1$ means the
next real token, skipping alignment padding. Its target loss mask determines
eligibility, allowing the last prompt token to predict the first assistant
token. Both objectives share one backbone pass and one output projection
over the union of selected positions, followed by one fused cross-entropy
computation and separate reductions.

\paragraph{Long-context sparse attention.}
The two streams double a 128K-token semantic sequence to approximately
256K stored positions. 
We describe permitted interactions with
structured metadata derived from example boundaries, stream roles,
semantic positions, and block identities. FlexAttention~\citep{dong2024flexattention} uses this metadata
to execute permitted attention tiles. The same identities guide BD-GDN,
convolution, and expert routing and are reused during activation
recomputation, keeping visibility consistent across the hybrid layers.

\paragraph{Parallelism for heterogeneous layers.}
Context parallelism shards a sequence across devices, but the required
communication differs by mixer. Full-attention layers keep local queries
and gather and reorder K/V. BD-GDN supports either a Ulysses
sequence-to-head transpose~\citep{jacobs2023ulysses} or recurrent-state
propagation using the context-parallel implementation in Flash Linear
Attention (FLA)\footnote{\url{https://github.com/fla-org/flash-linear-attention/blob/main/fla/ops/cp}}~\citep{yang2024fla}, which we refer
to as FLA-KCP.
Ulysses redistributes projected features before convolution and restores
the sequence-sharded layout before output projection; FLA-KCP preserves
ordered state dependencies across sequence shards. Expert parallelism
partitions routed experts across devices. Block metadata follows each
layout transformation, and sampled noise and block identities are kept
consistent across ranks and pipeline stages processing the same example.

\subsection{Decoding recipe}
\label{sec:decoding}
\paragraph{Expert routing.}
We use the block-level expert coreset described in Section~\ref{sec:dmoe}
for diffusion decoding or drafting, while causal verification retains token-level routing.

\paragraph{Confidence-aware decoding.} 
We decode the response block by block using a predict-and-commit strategy~\citep{wu2025fastdllmtrainingfreeaccelerationdiffusion,inclusionai2026llada22,cheng2026sdar,wu2026dreamreasoner}.
Given the committed prefix $x_{\le t}$, we append a block of $B$ mask
tokens and predict all $B$
masked positions in parallel. We commit the
predictions whose confidence exceeds a predefined threshold and remask
the remaining positions, which can be refined in the next diffusion step. Once all tokens are decoded, we update the clean memory and move to the next block.

\paragraph{Self-speculative decoding.}
Rather than committing tokens solely based on a fixed confidence
threshold, we use the causal GDN to verify the draft block generated
by a single diffusion step. We consider two acceptance rules:
\emph{exact verification} and \emph{entropy-based verification}. For each
draft token $d$, let $p$ denote its corresponding causal verification
distribution. Exact verification accepts $d$ only when it matches the
causal encoder's greedy prediction~\citep{leviathan2023fast,fu2026nemotronlabsdiffusion,chen2026dflash}, i.e.,
$d=\arg\max_{v\in\mathcal{V}}p(v)$.
In contrast, entropy-based verification adapts the acceptance threshold
based on the uncertainty of prediction~\citep{hewitt-etal-2022-truncation,cai2024medusa}.
Specifically, let $H=-\sum_{v\in\mathcal{V}}p(v)\log p(v)$ denote the
entropy of $p$. The draft token is accepted if
\begin{equation}
p(d)\geq\delta\exp(-H),
\end{equation}
where $\delta$ is a predefined hyperparameter. Under both rules, draft
tokens are verified sequentially until the first rejection, and only the
accepted prefix of the draft block is committed.
All subsequent evaluations of \clover use entropy-based self-speculative
decoding.

\paragraph{Decoding penalties.}
To mitigate repetition occasionally observed in diffusion decoding, we apply history-aware presence and repetition penalties to both confidence-aware and self-speculative decoding, with coefficients of $0.9$ and $1.05$, respectively. We additionally use length-dependent positive logit biases for stop tokens and the \texttt{</think>} token in thinking mode to encourage response termination and the transition from reasoning to the final answer.

\subsection{Benchmark performance}
\label{sec:experiments}
We evaluate Clover against LLaDA2.2 flash (104B-A7.4B), DiffusionGemma
(26B-A4B), Mercury 2.5, and the autoregressive Qwen3.6-35B-A3B and
Qwen3.5-35B-A3B models~\citep{inclusionai2026llada22,team2026diffusiongemmatechnicalreport,inceptionlabs2025mercury,qwen36_35b_a3b,qwen3.5}. Table~\ref{tab:main-results} covers 18 benchmarks
spanning knowledge, mathematics, code, reasoning, instruction following,
long-context understanding, and agent tasks.

\paragraph{Evaluation protocol.}
All main-table baselines are reproduced under the same evaluation protocol,
using each model's officially recommended decoding configuration. LLaDA2.2
flash uses its instruct model without a thinking mode; Mercury 2.5 uses
thinking at its default medium effort. Coding-agent evaluations use Terminus 2 for
Terminal-Bench~\citep{merrill2026terminalbench} and Codex as harness for SWE-bench~\citep{chowdhury2024swebenchverified}, with each task evaluated three times
and scores averaged. GDPval-AA v2~\citep{patwardhan2025gdpvalevaluatingaimodel} is evaluated locally using the recipe
provided by NVIDIA NeMo Gym~\citep{nemo-gym}. Terminal-Hard contains 21 hard tasks selected
from Terminal-Bench Core v0.1.1. Other benchmarks are evaluated using EvalScope~\citep{evalscope_2024}.

\clearpage
\begin{table}[p]
\centering
\caption{Main benchmark results. All models are evaluated in-house under the
protocol described in the text. Higher is better; \first{red bold} marks the
best value and \second{blue bold} the second-best distinct value within each
row, including ties. Shading identifies Clover. GDPval-AA v2 uses Elo, so its
differences are not percentage points.}
\label{tab:main-results}
\small\setlength{\tabcolsep}{4pt}
\renewcommand{\arraystretch}{1.22}
\arrayrulecolor{dsPrimary!45}
\begin{tabularx}{\linewidth}{L >{\columncolor{dsHighlight}}c c c c c c}
\toprule
\rowcolor{dsPrimary!8}
\textbf{Benchmark} & \makecell{\footnotesize\ours{Clover}\\\scriptsize 35B-A3B-Preview} & \makecell{\footnotesize\textbf{LLaDA2.2 flash}\\\scriptsize 104B-A7.4B} & \makecell{\footnotesize\textbf{DiffusionGemma}\\\scriptsize 26B-A4B} & \makecell{\footnotesize\textbf{Mercury 2.5}\\\scriptsize\strut} & \makecell{\footnotesize\textbf{Qwen3.6}\\\scriptsize 35B-A3B} & \makecell{\footnotesize\textbf{Qwen3.5}\\\scriptsize 35B-A3B}\\
\midrule
\rowcolor{dsPrimary!4}\multicolumn{7}{l}{\footnotesize\textcolor{dsPrimary}{\textbf{Knowledge}}}\\
GPQA Diamond & \first{83.8} & 47.5 & 68.2 & 76.3 & \second{83.3} & 78.3\\
HLE text-only & 12.8 & 5.3 & 10.9 & 8.9 & \second{21.2} & \first{21.6}\\
MMLU-Pro & \second{84.8} & 71.0 & 80.9 & 84.1 & 84.7 & \first{85.2} \\
\addlinespace[4pt]
\rowcolor{dsPrimary!4}\multicolumn{7}{l}{\footnotesize\textcolor{dsPrimary}{\textbf{Math}}}\\
MATH-500 & \second{96.2} & 93.2 & 94.4 & 93.8 & \second{96.2} & \first{97.2}\\
AIME 2026 (avg@8) & \second{83.3} & 56.7 & 68.3 & 48.8 & \first{90.4} & \first{90.4}\\
\addlinespace[4pt]
\rowcolor{dsPrimary!4}\multicolumn{7}{l}{\footnotesize\textcolor{dsPrimary}{\textbf{Code}}}\\
LiveCodeBench v6 & \second{72.6} & 39.4 & 63.4 & 64.0 & 71.4 & \first{74.3}\\
SciCode & \second{30.9} & 14.4 & 29.9 & 30.2 & \first{31.3} & 25.1\\
\addlinespace[4pt]
\rowcolor{dsPrimary!4}\multicolumn{7}{l}{\footnotesize\textcolor{dsPrimary}{\textbf{Reasoning}}}\\
ZebraLogic & \first{95.8} & 59.1 & 80.3 & 73.3 & 78.6 & \second{89.2}\\
\addlinespace[4pt]
\rowcolor{dsPrimary!4}\multicolumn{7}{l}{\footnotesize\textcolor{dsPrimary}{\textbf{Instruction following}}}\\
IFBench & \first{72.3} & 31.7 & 68.3 & \second{71.0} & 61.3 & 67.7\\
\addlinespace[4pt]
\rowcolor{dsPrimary!4}\multicolumn{7}{l}{\footnotesize\textcolor{dsPrimary}{\textbf{Long context}}}\\
AA-LCR & \first{72.0} & 33.0 & 47.0 & \second{70.0} & 66.0 & 57.0\\
\addlinespace[4pt]
\rowcolor{dsPrimary!4}\multicolumn{7}{l}{\footnotesize\textcolor{dsPrimary}{\textbf{Tool agent}}}\\
$\tau^3$ Overall & \first{65.3} & 51.4 & 32.1 & 58.0 & \second{60.9} & 44.3\\
BFCL-v4 & \first{67.9} & 59.3 & 63.4 & \second{66.1} & 63.4 & 62.2\\
\addlinespace[4pt]
\rowcolor{dsPrimary!4}\multicolumn{7}{l}{\footnotesize\textcolor{dsPrimary}{\textbf{Work agent}}}\\
GDPval-AA v2 (Elo) & \first{1091} & 651.0 & 721.0 & 742.0 & \second{1049.0} & 800.0\\
\addlinespace[4pt]
\rowcolor{dsPrimary!4}\multicolumn{7}{l}{\footnotesize\textcolor{dsPrimary}{\textbf{Coding agent}}}\\
Terminal-Bench 2.1 & \first{51.7} & 18.0 & 13.5 & 33.0 & \second{43.1} & 37.1\\
Terminal-Hard & \first{38.1} & 22.2 & 7.9 & 25.4 & \second{31.8} & 25.4\\
SWE-bench Verified & \first{73.6} & 49.3 & 23.8 & 57.8 & \second{65.8} & 61.4\\
SWE-Pro & \first{46.5} & 30.1 & 7.6 & 40.0 & \second{42.7} & 40.1\\
\bottomrule
\end{tabularx}
\arrayrulecolor{black}
\vspace{5pt}
\begin{minipage}{\linewidth}
\footnotesize
\textit{Model configuration.} LLaDA2.2 flash is available as an instruct model
without a thinking mode; Mercury 2.5 is evaluated with thinking enabled at
medium effort. Qwen3.6 and Qwen3.5 are autoregressive baselines.
\textit{Evaluation.} Coding-agent tasks use Terminus 2 for Terminal-Bench and
Codex for SWE-bench, averaging three evaluations per task. GDPval-AA v2 is
run locally using the NVIDIA NeMo Gym recipe. Terminal-Hard consists of 21
hard tasks selected from Terminal-Bench Core v0.1.1. Other benchmarks use
EvalScope. Baselines share the evaluation protocol and use each model's
officially recommended decoding configuration.
\end{minipage}
\end{table}

\clearpage

\paragraph{Agentic capability.}
Clover's clearest strength is its ability to turn reasoning into effective
actions, with a consistent lead across the seven evaluated agent benchmarks~\citep{barres2025tau2,patil2025bfcl,patwardhan2025gdpvalevaluatingaimodel,merrill2026terminalbench,chowdhury2024swebenchverified,deng2025swebenchproaiagents}.
SWE-bench Verified and Terminal-Bench are particularly informative because
they assess task completion in software repositories and executable
environments, where useful progress requires understanding context, choosing
actions, and responding to execution feedback. Strong performance in these
settings demonstrates capability beyond generating plausible code: the model
must help bring a task to a verifiable outcome. The accompanying results on
tool-use and professional-work evaluations broaden this picture, showing that
Clover's agent performance extends across different task types under the
reported evaluation setups.

\paragraph{General capabilities.}
The general benchmarks show that Clover's agent strengths are accompanied
by a broad base of knowledge, reasoning, and instruction-following capability.
ZebraLogic, IFBench, and AA-LCR are especially relevant~\citep{pyatkin2025ifbench,lin2025zebralogic,artificialanalysis2025lcr}: they test reasoning
under interdependent constraints, adherence to instructions, and the ability
to connect information across long documents. Clover's strong results on
these tasks are encouraging for applications that require both careful
interpretation of a task and sustained use of its context. Its competitive
knowledge-benchmark~\citep{rein2024gpqa,2026hle} performance further broadens this capability profile.
More demanding knowledge questions, competition mathematics, and standalone
coding remain useful directions for improvement. Overall, the results position
Clover as a capable general-purpose model with particular strength in agentic
applications.

\begin{takeaway}
Clover combines strong task execution across coding, tool-use, and
professional-work evaluations with competitive general capabilities.
Its strengths in reasoning, instruction following, and long-context
understanding complement this agentic capability profile.
\end{takeaway}

\subsection{Generation speed of self-speculative decoding}
\label{sec:speed}
\paragraph{Evaluation setup and workloads.}
We compare \clover with Qwen3.6-35B-A3B accelerated by DFlash 2~\citep{incoai2026dflash2} on a
single NVIDIA H200 GPU. All experiments use BF16 precision, a batch size of one, a block size of 16 and greedy decoding. Clover uses entropy-based self-speculative decoding with $\delta=0.6$, defined in Section~\ref{sec:decoding}. We select three representative workloads for evaluation: MATH-500, agent replay, and daily chat.

The agent replay corpus contains 64 prefixes from historical trajectories generated by Qwen3.6-35B-A3B. We select one
prefix per trajectory in sorted path order, alternating targets of 8K and
16K input tokens. Eligible prefixes end in a user message, contain 8K--24K
historically recorded input tokens, and precede a historical response of at
least 1,024 tokens. Thus, corpus construction is deterministic and targets
long-context generation. At evaluation time, the model generates a new
continuation; commands are not executed. Replay measures inference speed,
not a complete agent rollout or task success. 
The daily chat prompts cover everyday and explanatory topics, including
diffusion, sleep, memory, weather, batteries, refrigerators, the
Internet, GPS, seasons, bread, music, language learning, habits, stars, plants,
and photography. Each prompt requests a 1,200--1,600-word English response,
without imposing a minimum token count.

\paragraph{Measurement definitions.}
For request $i$, let $C_i$ be the completion-token count, $F_i$ the token
count in its first streamed event, $L_i$ its elapsed latency, $T_i$ its
time to first token (TTFT), and $V_i$ its number of verification rounds.
We report aggregate committed tokens per round and decode throughput as
\begin{equation}
 A=\frac{\sum_i C_i}{\sum_i V_i},\qquad
 \mathrm{TPS}_{\mathrm{decode}}
 =\frac{\sum_i(C_i-F_i)}{\sum_i(L_i-T_i)}.
 \label{eq:throughput}
\end{equation}
The accounting for $A$ includes bonus and boundary tokens and the first token
emitted by prefill. Decode TPS excludes prefill and
the wait before the first event, but includes thinking, answer text, and
client-side streaming overhead. It is a ratio of cumulative tokens to
cumulative time, not an average of per-request rates. We do not separately
report TTFT against autoregressive decoding because Clover uses the same
prefill computation; TTFT is therefore largely unchanged, and we focus on the decoding phase.

Table~\ref{tab:server-speed} shows that Clover sustains high committed-token
yields across all three workloads, exceeding DFlash 2 in tokens committed per
verification round. On MATH-500, however, its higher drafting cost offsets
this advantage, resulting in lower decode throughput than DFlash 2. On agent
replay and daily chat, Clover achieves both higher committed-token yields and
higher throughput. We attribute this advantage to Clover's drafting capability developing alongside the model throughout training. In contrast, methods such as DFlash train a separate lightweight drafter on target-generated responses over a selected prompt distribution, which may not cover the full range of workloads supported by the target. Expanding training data can improve coverage, but the drafter's limited capacity may still constrain how well it captures the target's behavior across workloads. 
These results highlight the importance of both draft cost and workload coverage when translating speculative progress into speed.

\begin{table}[htbp]
\centering
\begin{threeparttable}
\caption{H200 performance on three workloads. $A$ is committed tokens per
verification round; Decode TPS is defined in Eq.~\ref{eq:throughput}.}
\label{tab:server-speed}
\small\setlength{\tabcolsep}{5pt}\renewcommand{\arraystretch}{1.22}
\begin{tabular}{l l r r}
\toprule
\rowcolor{dsPrimary!8}\textbf{Workload} & \textbf{Model} & \textbf{$A$ (tokens/round)} & \textbf{Decode TPS}\\
\midrule
\rowcolor{dsPrimary!5}\cellcolor{white} & Clover-35B-A3B-Preview & \textbf{6.8743} & 465.2\\
\multirow{-2}{*}{\textbf{MATH-500}} & Qwen3.6-35B-A3B + DFlash 2 & 6.5897 & \textbf{604.3}\\
\cmidrule(lr){1-4}
\rowcolor{dsPrimary!5}\cellcolor{white} & Clover-35B-A3B-Preview & \textbf{8.2096} & \textbf{556.6}\\
\multirow{-2}{*}{\textbf{Agent replay}} & Qwen3.6-35B-A3B + DFlash 2 & 4.1548 & 396.7\\
\cmidrule(lr){1-4}
\rowcolor{dsPrimary!5}\cellcolor{white} & Clover-35B-A3B-Preview & \textbf{4.9970} & \textbf{356.7}\\
\multirow{-2}{*}{\textbf{Daily chat}} & Qwen3.6-35B-A3B + DFlash 2 & 2.9500 & 286.5\\
\bottomrule
\end{tabular}
\begin{tablenotes}[flushleft]\footnotesize
\item \textit{Notes.} Each configuration uses a fixed random samples.
Clover uses loose acceptance with coreset enabled. Decode TPS excludes prefill
and includes thinking. Bold marks the higher value for each metric within
each workload.
\end{tablenotes}
\end{threeparttable}
\end{table}

\begin{figure}[htbp]
 \centering
 \includegraphics[width=\linewidth]{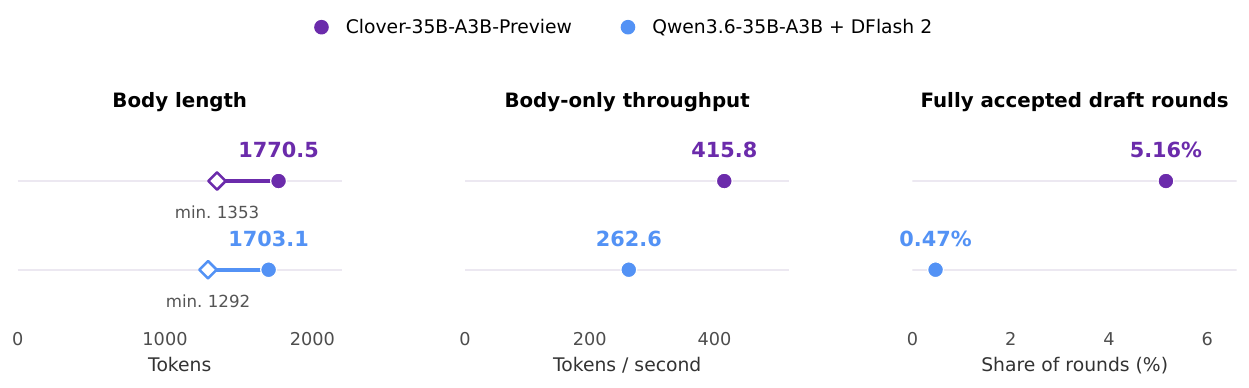}
 \caption{Daily chat: answer length, body-only decode throughput,
 and the fraction of fully accepted draft rounds. Filled circles show each
 metric; in the first panel, they indicate mean body length, while open diamonds
 indicate the minimum. Connecting segments span minimum to mean.
 Thinking is excluded from body-only TPS.}
 \label{fig:conversation-speed}
\end{figure}

\paragraph{Answer-phase behavior in daily chat.}
Figure~\ref{fig:conversation-speed} separates answer delivery from thinking.
Body-only decode TPS counts the remaining body tokens after the first body
event and divides by the remaining body-generation time; thinking tokens and
time are excluded. Both systems produce substantial answer bodies, while
Clover has higher body-only throughput and a larger fraction of rounds in
which every draft token is accepted. These body-only measurements complement the full-completion results in
Table~\ref{tab:server-speed}; the two metrics cover different output phases
and should be interpreted separately. 

\begin{figure}[!htbp]
\centering
\includegraphics[width=\linewidth]{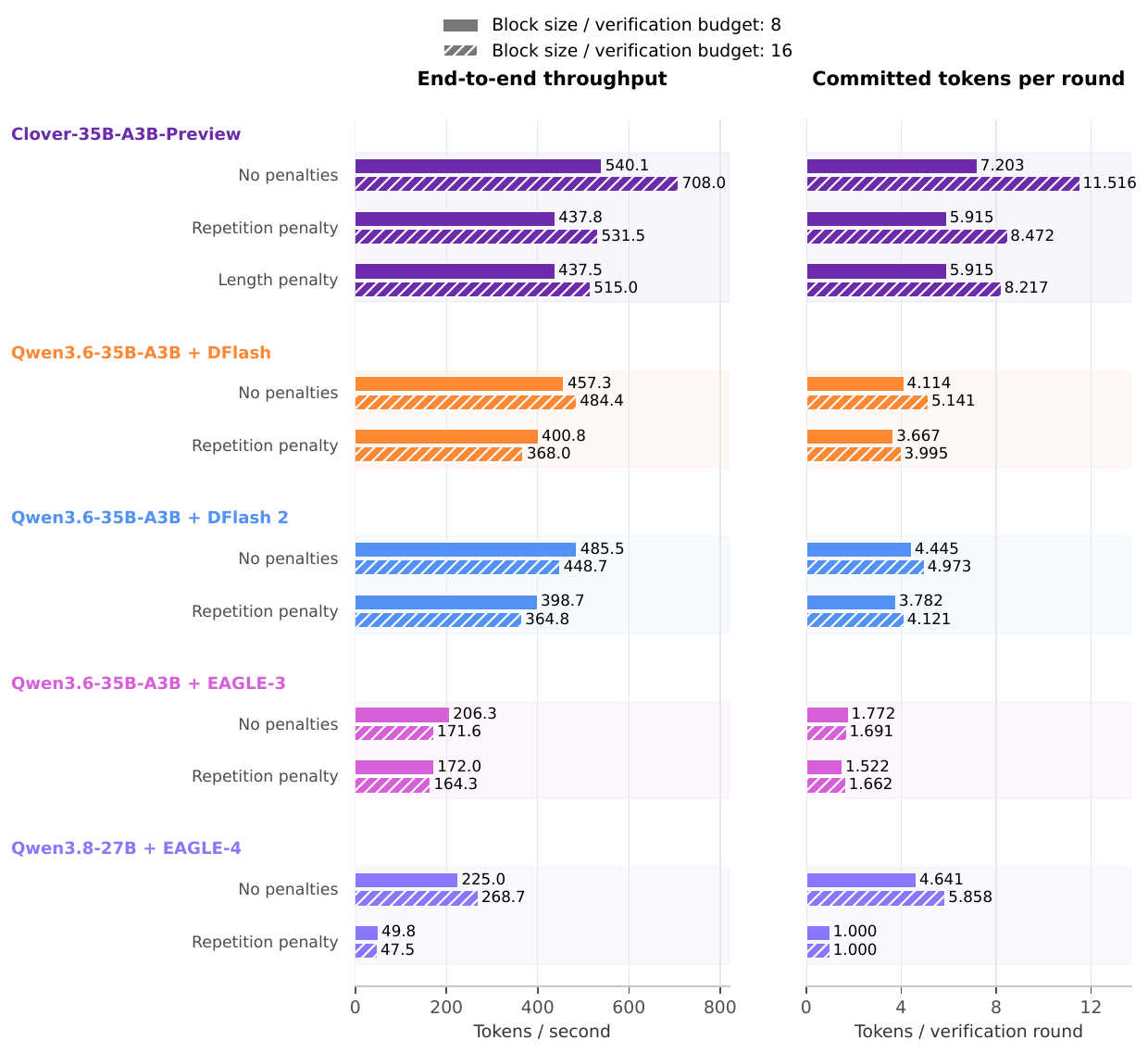}
\caption{Full agent replay: end-to-end throughput and committed tokens per
verification round. Solid/hatched bars use block size or verification budget
8/16. Clover is faster without penalties, but can sometimes produce repetitive
outputs.}
\label{fig:agent-replay-speed}
\vspace{2pt}
\begin{minipage}{\linewidth}\footnotesize
\textit{Notes.} Concurrency is capped at one. Each
setting uses natural EOS and a 16,384-token output cap. End-to-end timing includes prefill and tail drainage.

\emph{No penalties}: presence/repetition 0/1; \emph{Repetition penalty}:
0.9/1.05. Clover's \emph{Length penalty} additionally uses length penalty 10
(start/window 8192/8192) and thinking bias 10 (4096/4096).

EAGLE-3 uses the Caption drafter, depth 3, top-$k$ 4, and tree budget 8/16.
EAGLE-4 uses an adaptive tree within budget and native conservative acceptance
under penalties; its native decode rounds are included. DFlash 2 at block
size 16 is experimental.

Task success is not evaluated in this replay experiment.
\end{minipage}
\end{figure}

\paragraph{Sensitivity to penalties and verification budget.}
Figure~\ref{fig:agent-replay-speed} reports a separate sweep over all 64 agent replay
prefixes, comparing Clover with Qwen3.6-35B-A3B using DFlash, DFlash 2,
or EAGLE-3, and Qwen3.8-27B using EAGLE-4~\citep{chen2026dflash,incoai2026dflash2,li2025eagle3,eagle4team2026eagle4}.
This sweep measures end-to-end throughput, including prefill and tail drainage,
so its rates are not interchangeable with Eq.~\ref{eq:throughput}.
On this full 64-prefix agent replay test, Clover achieves 515 tokens/s on one
NVIDIA H200 GPU with block size 16 and the length and thinking controls enabled,
and reaches 680 tokens/s on one NVIDIA GB300 GPU with the same setting.
Clover has the highest throughput and committed-token yield among the plotted
systems under the shared penalty conditions at either budget. Increasing its
block size from 8 to 16 improves both quantities. For several other
configurations, a larger budget increases committed yield while reducing
throughput. Penalties also change the speed profile substantially, making
configuration disclosure essential. These are system-level comparisons:
different target models, verification trees, and decoding policies prevent
isolating the effect of the drafting algorithm, and task success is not
measured in replay.

\begin{takeaway}
Committed tokens per round quantify progress, but throughput also depends on
the cost of producing and verifying those tokens. Clover's self-speculative decoding extends acceleration across a broader
range of workloads, while further gains require continued optimization of
draft cost and decoding controls.
\end{takeaway}

\FloatBarrier

\clearpage
\section{On-Device Deployment}
\label{sec:edge}
\subsection{Comparison with efficient and on-device models}
Table~\ref{tab:edge-quality} compares Clover with MiniCPM-V 4.7 (35B-A3B),
LFM2.5 (24B-A2B), Nemotron 3.5 Lightning (30B-A3B), and Gemma 4 (26B-A4B)~\citep{openbmb2026minicpmv47,liquidai2026lfm25_24ba2b_modelcard,nvidia2026nemotron35lightning_modelcard,gemmateam2026gemma4}.
Clover and MiniCPM-V 4.7 are evaluated in-house under identical settings.
For the other models, the table combines official reported results, with
source-specific exceptions and harness differences retained in the notes.
This capability comparison is separate from the INT4 deployment measurement
in the next subsection.

\begin{table}[htbp]
\centering
\caption{Comparison with efficient and on-device models. Higher is better;
\first{red bold} marks the best reported value in each row, shading identifies
Clover, and dashes denote unavailable
results. Sources and harnesses differ as detailed below. These scores are
not asserted to come from Clover's INT4 deployment.}
\label{tab:edge-quality}
\small\setlength{\tabcolsep}{4pt}
\renewcommand{\arraystretch}{1.3}
\arrayrulecolor{dsPrimary!45}
\setlength{\benchmarkcolunit}{\dimexpr(.74\linewidth-12\tabcolsep)/5\relax}
\begin{tabular}{p{.26\linewidth} >{\centering\arraybackslash\columncolor{dsHighlight}}p{1.1\benchmarkcolunit} >{\centering\arraybackslash}p{1.15\benchmarkcolunit} >{\centering\arraybackslash}p{.7\benchmarkcolunit} >{\centering\arraybackslash}p{1.3\benchmarkcolunit} >{\centering\arraybackslash}p{.75\benchmarkcolunit}}
\toprule
\rowcolor{dsPrimary!8}
\textbf{Benchmark} & \makecell{\footnotesize\ours{Clover}\\\scriptsize 35B-A3B-Preview} & \makecell{\footnotesize\textbf{MiniCPM-V 4.7}\\\scriptsize 35B-A3B} & \makecell{\footnotesize\textbf{LFM2.5}\\\scriptsize 24B-A2B} & \makecell{\footnotesize\textbf{Nemotron 3.5}\\\scriptsize Lightning (30B-A3B)} & \makecell{\footnotesize\textbf{Gemma 4}\\\scriptsize 26B-A4B}\\
\midrule
IFBench & 72.3 & 47.7 & 74.7 & 71.9 & \first{77.2}\textsuperscript{*}\\
GPQA Diamond & \first{83.8} & 81.3 & 74.4 & 75.4 & 82.3\\
MMLU-Pro & \first{84.8} & 83.5 & --- & 81.9 & 82.6\\
HLE text-only & 12.8 & 9.4 & 7.4 & 11.7 & \first{17.2}\\
LiveCodeBench v6 & 72.6 & 56.6 & --- & --- & \first{77.1}\\
\midrule
BFCL-v4\textsuperscript{1} & \first{67.9} & --- & 65.1 & 53.8 & 57.1\\
$\tau^3$ Banking & \first{14.4} & 2.1 & 8.0 & 9.3 & 14.0\textsuperscript{*}\\
Terminal-Bench 2.1 & \first{51.7} & 36.0 & --- & 24.6 & 37.2\textsuperscript{*}\\
SWE-bench Verified\textsuperscript{2} & \first{73.6} & 45.6 & 51.0 & 51.6 & 57.4\textsuperscript{*}\\
\bottomrule
\end{tabular}
\arrayrulecolor{black}
\vspace{5pt}
\begin{minipage}{\linewidth}
\footnotesize
\textit{Evaluation notes.} Clover and MiniCPM-V 4.7 are evaluated in-house
under identical settings. Coding-agent results are three-run means, using
Terminus 2 for Terminal-Bench 2.1 and Codex for SWE-bench Verified. Unless
noted, LFM2.5 and Nemotron 3.5 Lightning scores come from their official
reports; unmarked Gemma 4 scores come from Google's model report. External
results may use different evaluation settings.

\textsuperscript{1} BFCL-v4 baselines for LFM2.5, Nemotron 3.5 Lightning,
and Gemma 4 are taken from the LFM2.5 report~\citep{liquidai2026lfm25_24ba2b_modelcard}.

\textsuperscript{2} LFM2.5 uses OpenCode for SWE-bench Verified, whereas
our in-house runs use Codex. Nemotron 3.5 Lightning and Gemma 4 also use
different coding-agent harnesses, following NVIDIA's evaluation report~\citep{nvidia2026nemotron35lightning_modelcard}.

\textsuperscript{*} Starred Gemma 4 results come from NVIDIA's
Nemotron 3.5 Lightning evaluation report~\citep{nvidia2026nemotron35lightning_modelcard}.
\end{minipage}
\end{table}

\FloatBarrier

Agentic capability is important for on-device deployment: a local model that
can use tools and carry out multi-step tasks can support applications beyond
conversation, including personal workflow automation, local coding assistance,
and interactions with device-resident software. Clover's strongest results in
Table~\ref{tab:edge-quality} align with this use case, spanning function calling,
interactive tool use, terminal tasks, and repository-level software engineering.
The matched comparison with MiniCPM-V 4.7 is particularly strong on the two
coding-agent benchmarks, while competitive knowledge and reasoning performance
provides a broader foundation for these applications.

There is also room to improve instruction following, difficult knowledge tasks,
and standalone code generation. Evaluation settings and coding-agent harnesses vary across external reports,
as detailed in the table notes. We next examine quantized deployment on
consumer hardware.

\subsection{Quantized deployment and runtime}
\paragraph{Deployment setup.}
We evaluate all configurations on the full MATH-500 benchmark~\citep{hendrycks2021measuring} using a single
NVIDIA RTX 5070 Ti with 16\,GB of VRAM, an AMD Ryzen 7 9700X CPU, and 64\,GB
of host memory. Inference runs in SGLang with KTransformers for hybrid
GPU--CPU expert execution~\citep{zheng2024sglang,chen2025ktransformers}, using eight CPU threads, a 32K context,
single-request inference, temperature zero, and thinking enabled.
Table~\ref{tab:edge-runtime} reports accuracy after manual review and decode
throughput excluding prefill and including thinking tokens.
Clover's routed expert weights use INT4 with group size 32, while the main
non-expert modules remain in BF16. Each layer places 96 experts on the GPU
and 160 on the CPU, with AVX-512 for CPU execution. The baselines use INT4
with group size 128, a 128/128 split, and AVX-VNNI-256. Clover's finer
group size provides more granular quantization scales, and its smaller
GPU-resident expert set leaves room for block-16 draft/verification
intermediates and the 32K cache. We use self-speculative decoding with the
block-level draft coreset threshold $\tau=0.9$ defined in
Section~\ref{sec:dmoe}.

\begin{table}[htbp]
\centering
\caption{Full MATH-500 evaluation on one RTX 5070 Ti (16\,GB), using SGLang
with KTransformers. Accuracy is manually reviewed; decode throughput excludes
prefill and includes thinking tokens.}
\label{tab:edge-runtime}
\small
\setlength{\tabcolsep}{3pt}
\renewcommand{\arraystretch}{1.3}
\setlength{\benchmarkcolunit}{\dimexpr\linewidth-12\tabcolsep\relax}
\begin{tabular}{
 >{\raggedright\arraybackslash}p{.35\benchmarkcolunit}
 >{\centering\arraybackslash}p{.12\benchmarkcolunit}
 >{\centering\arraybackslash}p{.13\benchmarkcolunit}
 >{\centering\arraybackslash}p{.18\benchmarkcolunit}
 >{\raggedleft\arraybackslash}p{.10\benchmarkcolunit}
 >{\raggedleft\arraybackslash}p{.12\benchmarkcolunit}}
\toprule
\rowcolor{dsPrimary!8}
\textbf{Model / placement} & \makecell{\textbf{Expert}\\\textbf{weights}} &
\makecell{\textbf{GPU / CPU}\\\textbf{experts}\textsuperscript{1}} &
\makecell{\textbf{CPU}\\\textbf{backend}} &
\makecell{\textbf{Accuracy}\\\textbf{(\%)}} &
\makecell{\textbf{Decode}\\\textbf{(tokens/s)}}\\
\midrule
MiniCPM-V 4.7 & INT4 g128 & 128 / 128 & AVX-VNNI-256  & 97.4 & 49.51\\
Qwen3.6-35B-A3B & INT4 g128 & 128 / 128 & AVX-VNNI-256  & 93.8 & 47.04\\
\midrule
\multicolumn{6}{l}{\textcolor{dsPrimary}{\textbf{Clover-35B-A3B-Preview}}}\\
\addlinespace[2pt]
\mbox{Original placement, top-8} & INT4 g32 & 96 / 160 & AVX-512  & 96.8 & 66.64\\
\mbox{Optimized placement, top-8} & INT4 g32 & 96 / 160 & AVX-512  & \textbf{97.6} & 97.52\\
\highlightrow
\mbox{Optimized placement, top-6} & INT4 g32 & 96 / 160 & AVX-512  & 96.0 & \textbf{106.25}\\
\bottomrule
\end{tabular}
\vspace{4pt}
\begin{minipage}{\linewidth}\footnotesize
\textsuperscript{1} Resident experts per layer. Optimized placement changes
expert identities, not the GPU/CPU counts. Top-8 and top-6 denote active experts
per token during verification.
\end{minipage}
\end{table}

\paragraph{Routing-informed expert placement.}
Expert placement strongly affects hybrid execution because CPU expert
computation is slower and can dominate generation time. We use routing
statistics from mathematical tasks to place frequently used experts on the
GPU and less frequently used experts on the CPU, reducing CPU computation
without changing the number of experts resident on either device. With
top-eight verification, this improves decode throughput from 66.64 to
97.52 tokens/s, with 97.6\% reviewed accuracy. Reducing verification to six
active experts per token further increases throughput to 106.25 tokens/s
at 96.0\% accuracy, offering a faster operating point through approximate
computation. These results highlight the value of matching expert placement
to workload demand alongside decoding optimization.

\paragraph{Draft sparsity and acceptance.}
We also tested a lower draft coreset threshold of $\tau=0.6$. A smaller
coreset is intended to reduce the number of experts used during drafting
on both GPU and CPU. However, this setting also reduced the accepted draft
length during verification, and the resulting throughput was lower than
with $\tau=0.9$. Thus, reducing draft computation alone does not necessarily
improve speed: the saved work must be balanced against the number of tokens
committed per verification round.

\begin{takeaway}
Workload-aware expert placement and self-speculative decoding enable fast
Clover inference on a 16\,GB consumer GPU. Expert placement, draft sparsity,
and verification cost provide complementary controls for balancing generation
speed and model quality.
\end{takeaway}

\FloatBarrier
\FloatBarrier
\section{Clover Decision Model}
\label{sec:decision}

Human decision-making is often described in terms of fast, intuitive
System~1 thinking and slower, deliberate System~2 reasoning.
Drawing on this distinction, Jev, a commercial model from TypeSafe AI,
targets fast, structured decisions that software can use
directly~\citep{almeida2026jev}.
This motivates the Clover decision model: we use block diffusion to produce
related judgments jointly in a single denoising pass, combining fast response
with the ability to share information across decisions.

\subsection{Joint decisions in one denoising step}
The Clover decision model is a variant for fast, structured judgment tasks. Given shared material and a set of questions, it predicts up to eight
answers together in a single denoising pass after context prefill. This
variant is planned for a subsequent release.

Let $c$ denote the shared material and $q_1,\ldots,q_M$ the associated
questions, with $M\leq 8$. We place these in the input and provide a fixed
answer template in one diffusion block. Each question has one masked answer
position, while the surrounding format is given. One denoising pass predicts
all answer positions, and each prediction is read over that question's
candidate labels. Independent mode instead invokes the same model separately
for each $(c,q_j)$.

All answer positions share the input and can exchange intermediate features
within the block. This provides a way to use related evidence across
decisions: for example, several judgments about a contract may depend on the
same disclosure exception. The mechanism supports coordinated prediction,
although a single pass does not guarantee that the resulting answers satisfy
every constraint. 

\subsection{Decision quality}
\label{sec:decision-results}
We compare joint and independent prediction using the same decision-model
checkpoint on an internal set of 500 groups with 2,420 questions and a set of
50 partially completed $4\times4$ Sudoku boards with 364 missing entries.
Alongside accuracy over individual questions or cells, we report the fraction
of groups or boards for which every answer is correct.

\begin{table}[htbp]
\centering
\caption{Joint versus independent prediction with the same Clover decision
model. All values are percentages; higher is better. Complete-set accuracy
requires every question in a group or every missing cell in a board to be
correct.}
\label{tab:decision}
\begin{tabularx}{\linewidth}{L l r r}
\toprule
Task & Mode & Question / cell accuracy & Complete-set accuracy\\
\midrule
Multiple500 & Independent & 90.37 & 65.4\\
\highlightrow Multiple500 & Joint & \textbf{93.88} & \textbf{77.0}\\
\midrule
Sudoku & Independent & 78.30 & 22.0\\
\highlightrow Sudoku & Joint & \textbf{95.60} & \textbf{76.0}\\
\bottomrule
\end{tabularx}
\end{table}

Joint prediction improves complete-set accuracy by 11.6 percentage points on
Multiple500 and 54.0 points on Sudoku (Table~\ref{tab:decision}). The larger
board-level improvement is particularly relevant to tasks with coupled
constraints, where a useful result depends on getting the full set of
decisions right. These results support the value of joint prediction on the
evaluated tasks; complete-set accuracy alone does not establish the mechanism
behind the gain or directly measure logical consistency.

We also evaluate the same checkpoint on JevBench and the full Decision
Index~0.2.1 suite~\citep{standhartinger2026jevbench,multimodalart2026decisionindex}. Table~\ref{tab:decision-external} compares it with our frozen
Jev API reference. Decision Index retains each task's native metric and
aggregates normalized scores; its index is not an accuracy percentage.
These results assess broader decision capability rather than isolating the
benefit of joint prediction. The internal sets have been used repeatedly
during development, so they serve as capability checks rather than new blind
tests.

\begin{table}[htbp]
\centering
\caption{External decision benchmarks. JevBench reports accuracy over 231
questions; Decision Index reports the full suite's composite score. Both
metrics are higher-is-better. Jev v1.13 is a frozen API evaluation reference.}
\label{tab:decision-external}
\begin{tabularx}{\linewidth}{L r r}
\toprule
Model & JevBench accuracy (\%) & Decision Index 0.2.1\\
\midrule
Jev v1.13 & 86.58 & 57.93\\
\highlightrow Clover-Decision & \textbf{88.31} & \textbf{58.23}\\
\bottomrule
\end{tabularx}
\end{table}

\subsection{Efficiency of joint prediction}
\begin{figure}[htbp]
\centering
\includegraphics[width=.82\linewidth]{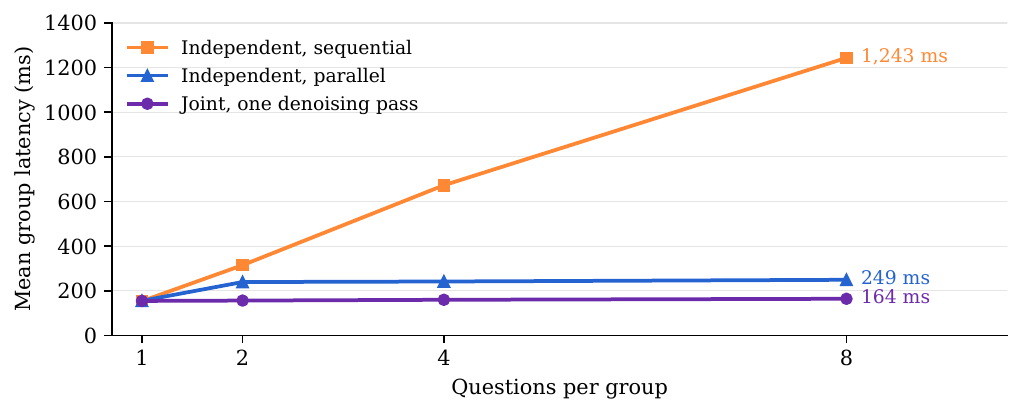}
\caption{\textbf{Group latency as the number of questions increases.}
Measurements use the same GPU and checkpoint, with one outer task at a time
and a single denoising pass per call. Each point averages 16 measurements: four repetitions on each of four fixed cases. We vary group size by selecting the first 1, 2, 4, or 8 questions from each case.}
\label{fig:decision-latency}
\end{figure}

Joint mode amortizes shared context processing and model invocation across
questions while reading all answers in one denoising pass. Figure~\ref{fig:decision-latency}
compares it with sequential and parallel independent calls under the same
hardware, checkpoint, and runtime configuration. With eight questions,
mean group latency is 164\,ms for joint prediction, compared with 1,243\,ms
for sequential independent calls and 249\,ms for parallel independent calls,
corresponding to speedups of $7.58\times$ and $1.52\times$, respectively.
The parallel comparison allows up to eight independent requests at once.
Even when independent requests are batched, each retains its own context and
answer block, whereas joint prediction processes the shared material once
and places all answers in one block. 

\begin{takeaway}
Predicting related answers together improves individual and complete-set
accuracy on the evaluated multi-question tasks. A single denoising pass also
reduces group latency relative to both sequential and parallel independent
calls, providing a practical use of block diffusion for structured decisions.
\end{takeaway}

\FloatBarrier
\section{Conclusion}
\label{sec:conclusion}
We presented \clover, a 35B-total, 3B-active mixture-of-experts diffusion language
model built around a bidirectional adaptation of Gated Delta Networks. Its
architecture combines recurrent historical memory, selective interactions
within the current block, learnable block-size calibration, and bidirectional
local features. The benchmark results show a consistent lead across the evaluated agent
metrics alongside competitive general capabilities, with opportunities for
further improvement in knowledge and mathematics.
Quantized deployment offers a further direction
for studying this architecture under memory constraints. Our decision-model
results additionally suggest that jointly generating related answers can
improve the correctness of a complete set of judgments.

The next priorities are to further reduce latency, characterize the
quality--latency trade-off under matched evaluation conditions, and strengthen
quantized deployment. We are also extending Clover toward multimodal
capabilities, informed by Dream-VL and Dream-VLA work on
vision-language understanding and robotic action modeling~\citep{ye2025dreamvldream}, and extending the context length.

\section*{Acknowledgments}
We thank the Qwen, SGLang, and llama.cpp communities for the open model and
inference infrastructure on which this work builds.
\section*{Contributors}
All authors are affiliated with DiffuSpace.
Contributors are listed alphabetically by surname, reading down each column.
Important contributors are shown in bold; $\dagger$ denotes corresponding authors.

We also thank the supporting teams across DiffuSpace, whose dedication and contributions beyond the technical work made this release possible.

\medskip
\noindent
\begin{tabularx}{\linewidth}{@{}L L@{}}
Chen, Jingdi & \textbf{Qi, Guicheng}\textsuperscript{\ensuremath{\dagger}}\\
Chen, Zizhao & Ruan, Jianhao\\
Deng, Shengliang & \textbf{Wu, Zirui}\textsuperscript{\ensuremath{\dagger}}\\
\textbf{Feng, Xiachong}\textsuperscript{\ensuremath{\dagger}} & Yao, Jinwei\\
\textbf{Gong, Shansan}\textsuperscript{\ensuremath{\dagger}} & \textbf{Ye, Jiacheng}\textsuperscript{\ensuremath{\dagger}}\\
Huang, Yujin & Zhang, Shiying\\
{Kong, Lingpeng} & \textbf{Zhang, Zongpu}\\
\textbf{Li, Mukai} & Zhu, Jiatong\\
Peng, Qi & \\
\end{tabularx}

\medskip
{\footnotesize
\noindent $\dagger$ \textit{Correspondence:}
Jiacheng Ye (\email{jcye@diffuspace.ai});
Shansan Gong (\email{sansa@diffuspace.ai});
Guicheng Qi (\email{gcqi@diffuspace.ai});
Zirui Wu (\email{ziruiwu@diffuspace.ai});
Xiachong Feng (\email{xcfeng@diffuspace.ai}).\par
}

\bibliographystyle{diffuspace}
\bibliography{references}
\end{document}